%% file: main.tex
\documentclass{article} %
\usepackage{openmoss}
\usepackage{microtype}
\definecolor{linkblue}{RGB}{0,0,139}      %
\definecolor{navy}{RGB}{0,0,128}          %
\definecolor{royalblue}{RGB}{65,105,225}  %
\definecolor{steelblue}{RGB}{70,130,180}  %
\definecolor{dodgerblue}{RGB}{30,144,255} %
\definecolor{mediumblue}{RGB}{0,0,205}    %
\definecolor{darkslateblue}{RGB}{72,61,139} %
\usepackage[colorlinks = true,
            linkcolor = linkblue,
            urlcolor  = royalblue,
            citecolor = royalblue,
            anchorcolor = blue]{hyperref}

\usepackage{amssymb}

\usepackage{url}
\usepackage{booktabs}
\usepackage{graphicx}
\usepackage{setspace}
\usepackage{enumitem}
\usepackage{times}
\usepackage{latexsym}

\usepackage[T1]{fontenc}

\usepackage[utf8]{inputenc}

\usepackage{microtype}

\usepackage{inconsolata}
\usepackage{fancyhdr}

\usepackage{graphicx}
\usepackage{subcaption}
\usepackage{multirow}  %
\usepackage{tcolorbox}

\usepackage{amsfonts}       %
\usepackage{nicefrac}       %
\usepackage{xcolor}         %
\usepackage[colorinlistoftodos]{todonotes}
\usepackage{amsmath}
\usepackage{algorithm}
\usepackage{algpseudocode}
\usepackage{multicol}
\usepackage[font=small,labelfont=bf,tableposition=top]{caption}
\usepackage{colortbl}
\usepackage{cleveref}
\usepackage{parskip}  %
\usepackage{calc}
\usepackage[utf8]{inputenc}
\usepackage[english]{babel}
\usepackage{tcolorbox}
\usepackage{fvextra}
\usepackage[utf8]{inputenc}
\usepackage[T1]{fontenc}   %
\usepackage{lipsum} 
\usepackage{booktabs}
\tcbuselibrary{breakable} %
\newtcolorbox{promptblock}{
  colback=blue!2!white,
  colframe=blue!30!gray,
  boxrule=0.6pt,
  arc=3pt,
  left=12pt,
  right=12pt,
  top=8pt,
  bottom=8pt,
  boxsep=0pt,
  before skip=10pt,
  after skip=10pt,
  breakable,
  fontupper=\normalfont,
  parbox=false,
}

\input{math_commands.tex}

\title{DiRL: An Efficient Post-Training Framework for Diffusion Language Models}

\colmfinalcopy

\author{
Ying Zhu$^{1,2,3}$, Jiaxin Wan$^{2}$, Xiaoran Liu$^{1,2,3}$, Siyang He$^{1,2,3}$, Qiqi Wang$^{1,2,3}$, \\[1ex]
\ \textbf{Xu Guo$^{1,2}$, Tianyi Liang$^{2,3}$, Zengfeng Huang$^{1,2}$, Ziwei He$^{2,3\dag}$, Xipeng Qiu$^{1,2\dag}$}
\\[1ex]
$^{1}$Fudan University, 
$^{2}$Shanghai Innovation Institute, 
$^{3}$OpenMoss Team \\[1ex]
$^{\dag}$Corresponding author.
}

\fancypagestyle{firstpage}{
  \lhead{\begin{picture}(0,0)\put(0,-6){\includegraphics[width=0.3\linewidth]{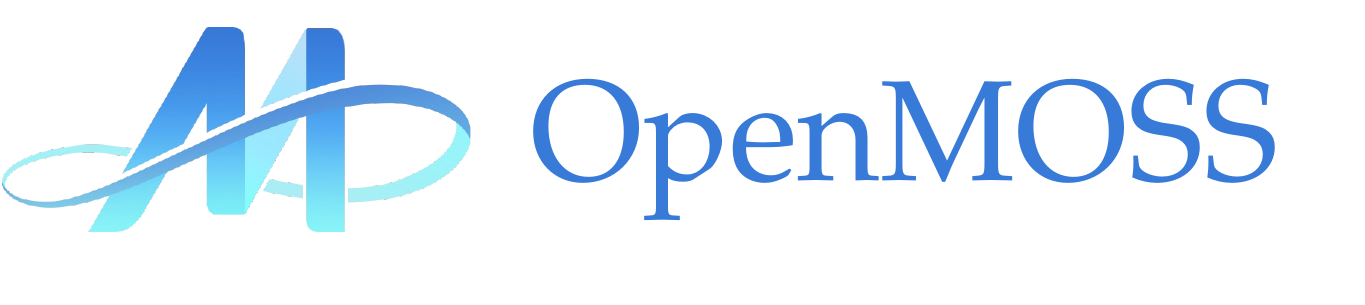}}\end{picture}}
}

\newcommand{\ctext}[1]{\raise0.2ex\hbox{\textcircled{\scriptsize{#1}}}}

\begin{document}

\maketitle
\thispagestyle{firstpage}

\newcounter{num}
\newcommand{\rnum}[1]{\setcounter{num}{#1} \roman{num}}
\crefname{equation}{Eq.}{Eqs.}

\begin{abstract}
Diffusion Language Models (dLLMs) have emerged as promising alternatives to Auto-Regressive (AR) models. While recent efforts have validated their pre-training potential and accelerated inference speeds, the post-training landscape for dLLMs remains underdeveloped. Existing methods suffer from computational inefficiency and objective mismatches between training and inference, severely limiting performance on complex reasoning tasks such as mathematics. To address this, we introduce DiRL, an efficient post-training framework that tightly integrates FlexAttention-accelerated blockwise training with LMDeploy-optimized inference. This architecture enables a streamlined online model update loop, facilitating efficient two-stage post-training (Supervised Fine-Tuning followed by Reinforcement Learning). Building on this framework, we propose DiPO, the first unbiased Group Relative Policy Optimization (GRPO) implementation tailored for dLLMs. We validate our approach by training DiRL-8B-Instruct on high-quality math data. Our model achieves state-of-the-art math performance among dLLMs and surpasses comparable models in the Qwen2.5 series on several benchmarks.

{\fontsize{10pt}{10pt} \selectfont \raisebox{-0.06em}{\includegraphics[height=1em]{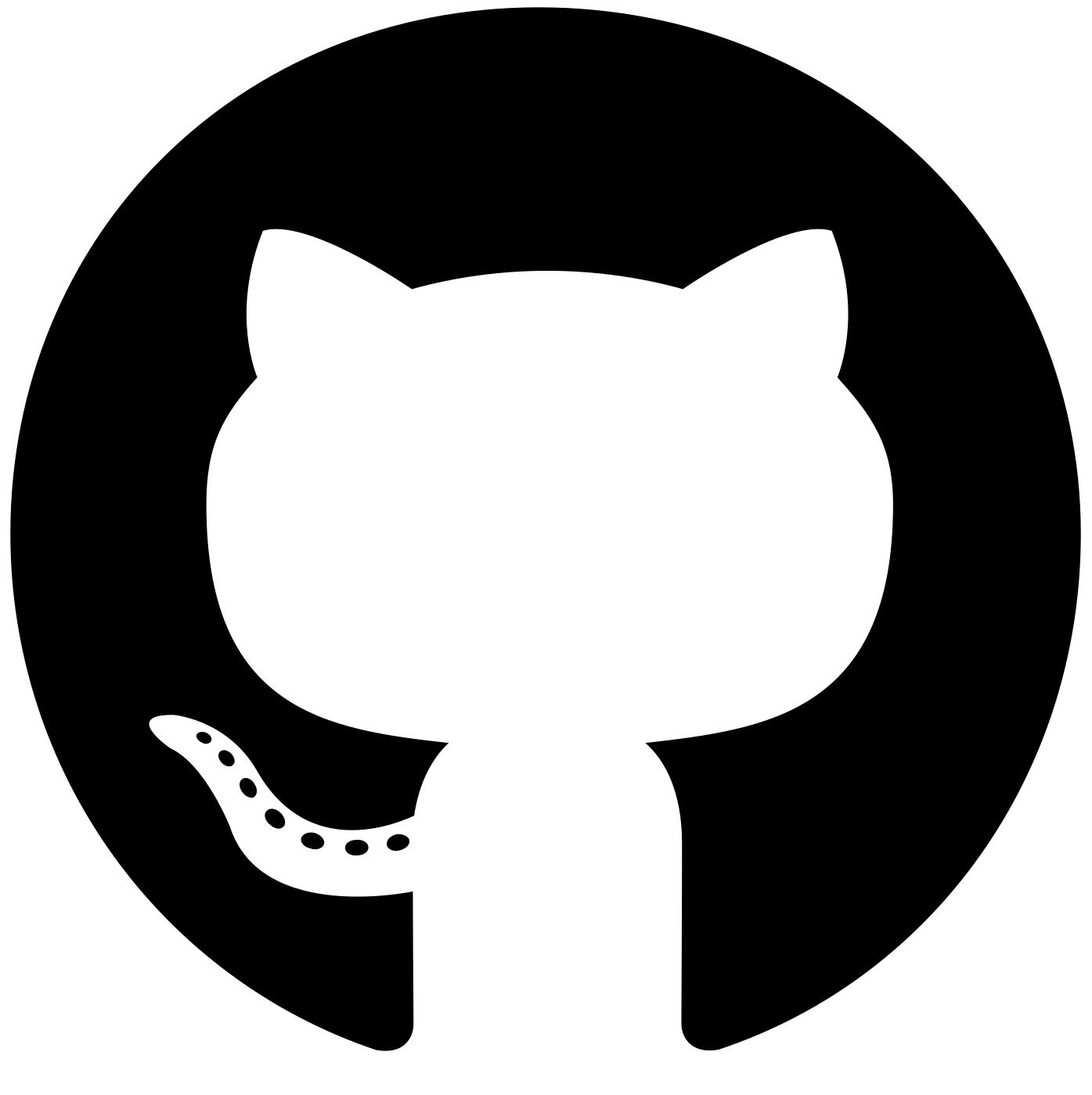}} Code: \href{https://github.com/OpenMOSS/DiRL}{https://github.com/OpenMOSS/DiRL} \\
\raisebox{-0.06em}{\includegraphics[height=1em]{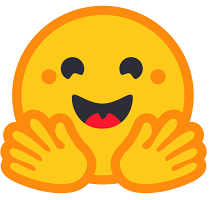}} Model: \href{https://huggingface.co/OpenMOSS-Team/DiRL-8B-Instruct}{https://huggingface.co/OpenMOSS-Team/DiRL-8B-Instruct}}
\end{abstract}

\section{Introduction}

\begin{figure}[!b]
    \begin{minipage}[b]{0.41\linewidth}
        \centering
        \includegraphics[width=\linewidth]{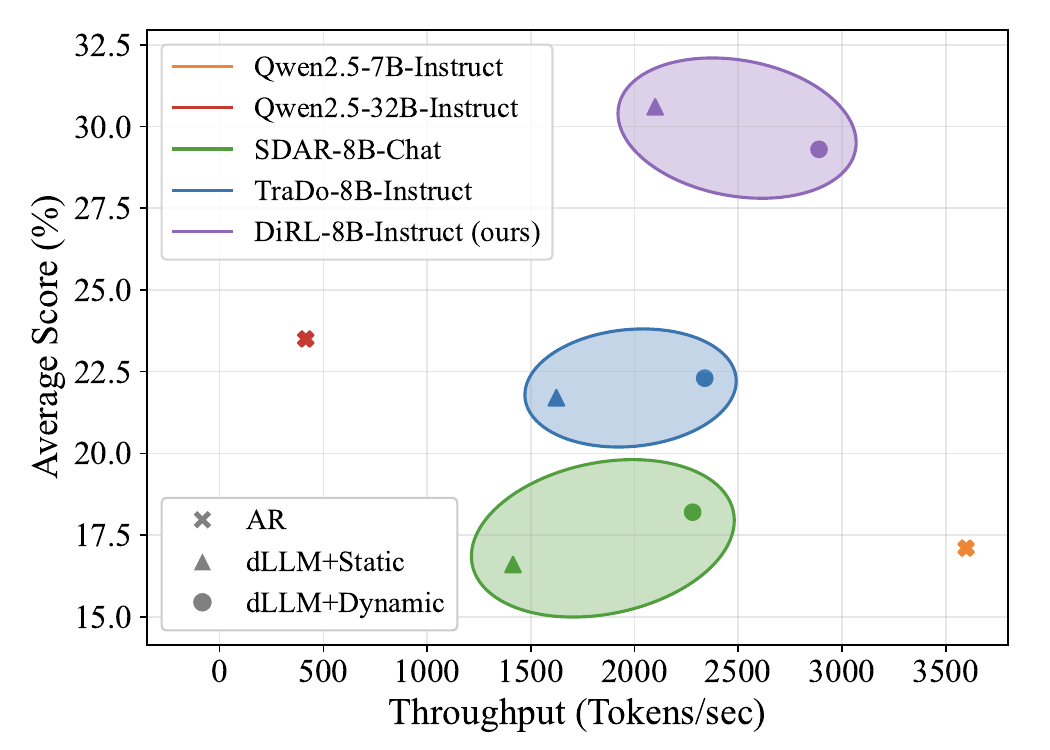}
        \caption{Performance of DiRL-8B-Instruct.}
        \label{dirl_performance}
    \end{minipage}
    \hfill
    \begin{minipage}[b]{0.57\linewidth}
        \centering
        \includegraphics[width=\linewidth]{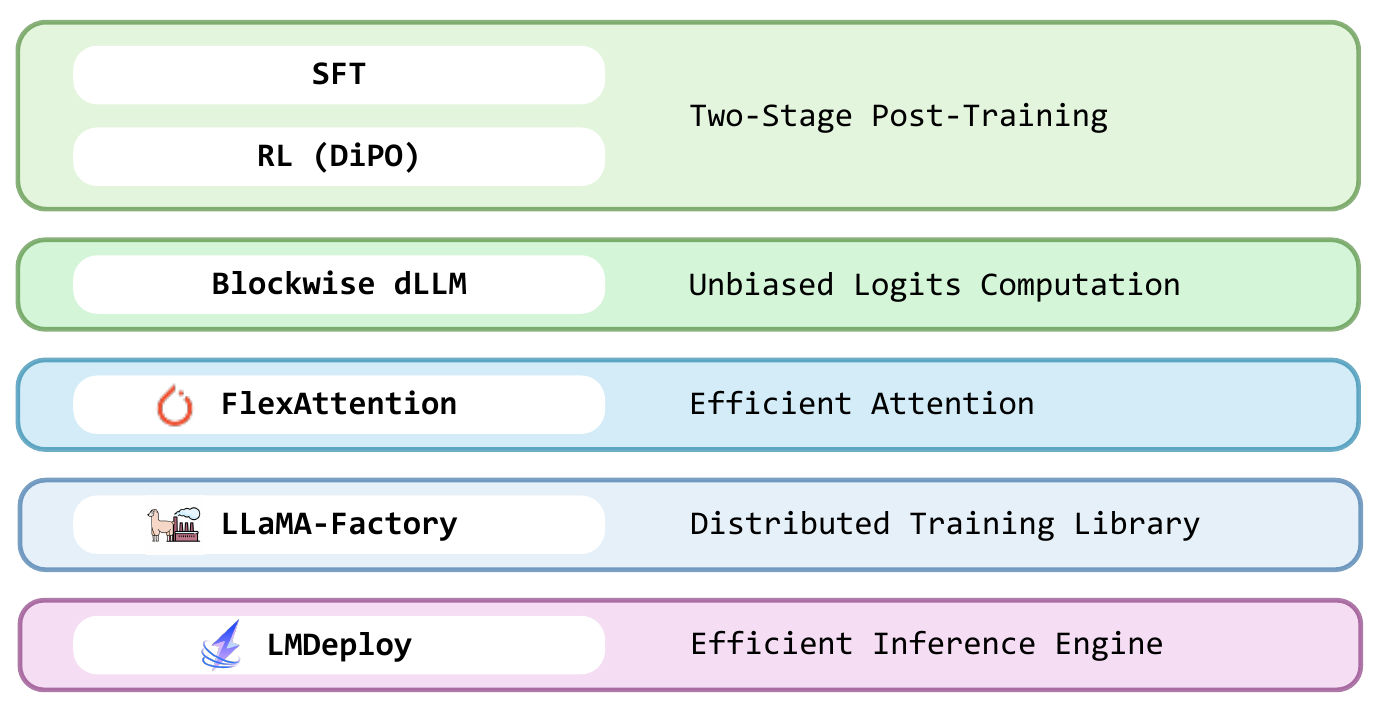}
        \caption{Features of our DiRL framework.}
        \label{dirl_framework}
    \end{minipage}
\end{figure}

Large Diffusion Language Models (dLLMs) have become a hot topic in NLP~\citep{nie2025large,mercury}. The emergence of dLLMs such as LLaDA~\citep{nie2025large}, Dream~\citep{ye2025dream}, Mercury~\citep{mercury} and Gemini Diffusion~\citep{genimid}, together with blockwise hybrids like SDAR~\citep{cheng2025sdar} that combine diffusion with traditional auto-regression (AR), confirms the scalability of this paradigm~\citep{nie2024scaling,gong2024scaling,ni2025training}. Building on these results, a growing body of work now seeks to advance their performance in multi-modality~\citep{yang2025mmada,you2025llada}, long-context modeling~\citep{liu2025longllada,he2025ultrallada} and inference efficiency~\citep{wu2025fast,wu2025fast2,song2025sparse}. Although pre-training of dLLMs is now proven feasible, post-training of dLLMs, especially reinforcement learning (RL), remains underdeveloped, limiting dLLMs' performance on math tasks and real-world deployment.

\begin{figure}[!t]
    \centering
    \includegraphics[width=0.96\linewidth]{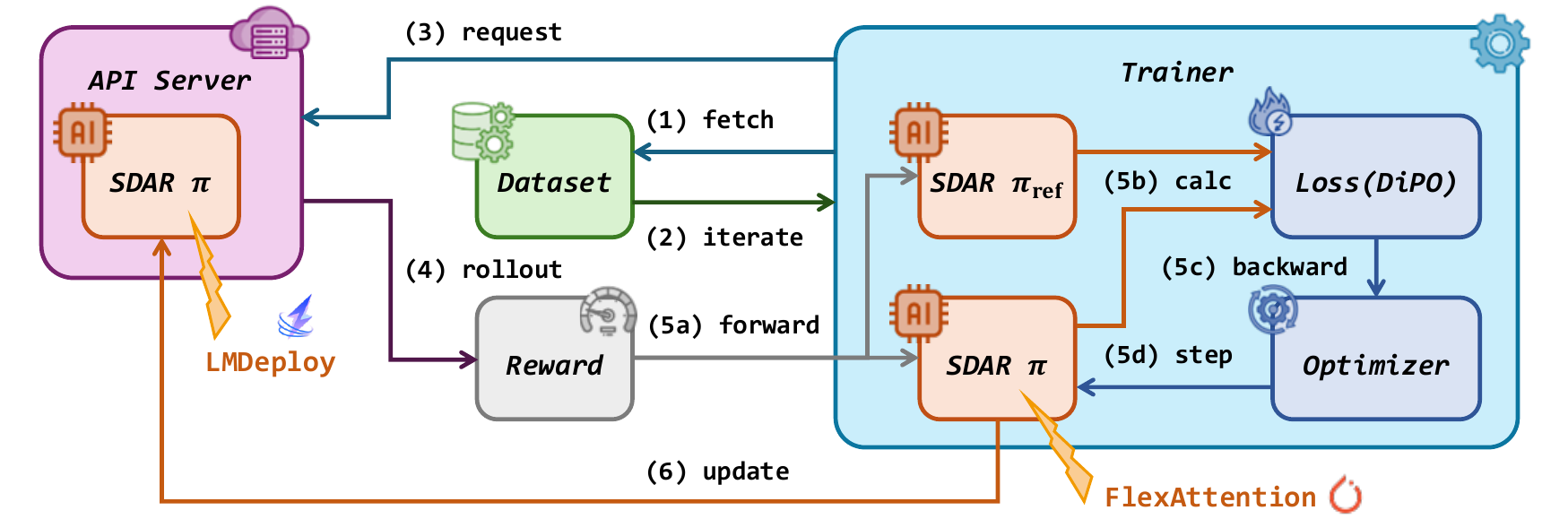}
    \caption{Overview of the RL pipeline in our DiRL framework.}
    \label{dirl_pipeline}
\end{figure}

The difficulty of dLLM post-training, especially RL, lies in the fact that the logits and derived policy cannot be computed exactly~\citep{zhao2025d1,zhu2025llada}. In original fully bidirectional dLLMs, the generation order is unconstrained, making teacher-forcing style logit acquisition during SFT infeasible. Injecting uniform random noise into the output fails to reproduce the realistic inference step map, resulting in biased logits and a large mismatch between training and inference objectives~\citep{zhao2025d1,wang2025revolutionizing}. Furthermore, in the RL stage, the absence of a KV cache further increases computational overhead~\citep{liudllm,ma2025dkv,song2025sparse}. Most existing dLLM-based RL efforts lack an inference-engine backend, efficient training–inference co-design, and fast rollouts with online model updates, which prevents the practical adoption of mature RL algorithms such as GRPO~\citep{deepseekv3,guo2025deepseek}. Blockwise dLLMs partially alleviate these issues by restricting generation within blocks, enabling exact logit computation through blockwise forward passes~\citep{cheng2025sdar,wang2025revolutionizing}. However, they do not fully resolve the train–inference mismatch in post-training, nor do they address the efficiency and algorithmic challenges of dLLM RL. How to achieve consistent training and inference while enabling scalable RL for dLLMs remains underexplored.
To fill this gap, we introduce our dLLM RL algorithm \textit{\textbf{DiPO}}, together with the training framework \textit{\textbf{DiRL}}, which enforces training–inference consistency and enables efficient rollouts and policy optimization for blockwise dLLMs. We further present the state-of-the-art dLLM \textit{\textbf{DiRL-8B-Instruct}}, as shown in Figure~\ref{dirl_performance}, Figure~\ref{dirl_framework} and Figure~\ref{dirl_pipeline}.

At the algorithm level, \textit{\textbf{DiPO}} leverages the good property of blockwise dLLM to achieve the first unbiased GRPO implementation for dLLMs through efficient, unbiased logit computation. At the framework level, \textit{\textbf{DiRL}} supports the two-stage (SFT-RL) post-training of dLLMs, aligning training and inference objectives while surpassing existing methods in efficiency as shown in Figure~\ref{dirl_framework}. Concretely, we integrate the efficient inference property of blockwise dLLMs with the efficient FlexAttention interface~\citep{dong2024flex} and LMDeploy framework~\citep{2023lmdeploy} to enable fast rollout and online model updates in the API server. At the model level, based on high-quality math datasets, we train \textit{\textbf{DiRL-8B-Instruct}} from SDAR-8B-Chat and achieve best performance on math tasks in dLLMs, even outperform Qwen2.5 Series~\citep{qwen2024qwen25technicalreport}, the widely-acknowledged larger AR model, in AIME24, AIME25~\citep{aime2024,aime2025} and OlympiadBench~\citep{he2024olympiadbench} as shown in Figure~\ref{dirl_performance}. Our contributions can be summarized as follows.
\begin{itemize}
    \item \textbf{\textit{DiRL}}, 
    an efficient post-training framework for dLLMs that replaces offline model loading with inference-server-based rollouts and online policy updates, ensuring training–inference consistency and accelerated by FlexAttention.    
    \item \textbf{\textit{DiPO}}, the first unbiased GRPO implementation in dLLMs, leveraging the unbiased logits computation of blockwise dLLM.
    \item \textbf{\textit{DiRL-8B-Instruct}}, the state-of-the-art dLLMs in math tasks, based on the above algorithm and engineering improvements, as well as high-quality math data.
\end{itemize}

\section{Preliminary}

\subsection{Block Diffusion Language Model}
Blockwise Diffusion Language Models, as exemplified by BD3-LMs~\citep{arriola2025block} and SDAR~\citep{cheng2025sdar}, primarily unify the global sequential dependency of AR models with the local parallel generation capability of dLLMs through a Semi-Autoregressive generation paradigm. 

In blockwise dLLMs, given a discrete sequence $\mathbf{x}$, we partition it into $K$ non-overlapping text blocks, denoted as $\mathbf{x} = (\mathbf{b}^1, \mathbf{b}^2, \dots, \mathbf{b}^K)$, where each block contains $B$ tokens. Then the joint probability distribution of the sequence $p_\theta(\mathbf{x})$ is factorized into a product of conditional probabilities, where $\mathbf{b}^{<k}$ represents the historical context preceding the current block.
\begin{equation}
\log p_\theta(\mathbf{x}) = \sum_{k=1}^{K} \log p_\theta(\mathbf{b}^k \mid \mathbf{b}^{<k})
\end{equation}

In contrast to the token-by-token generation of AR, the intra-block conditional distribution $p_\theta(\mathbf{b}^k \mid \mathbf{b}^{<k})$ is modeled by a discrete masked diffusion process conditioned on the historical information.

\paragraph{Forward Process (Noising Process)}
A Markov chain $q(\mathbf{b}^k_t \mid \mathbf{b}^k_0)$ is defined, where as the time step $t \in [0, 1]$ increases, tokens in the current block $\mathbf{b}^k_0$ are progressively replaced by the \texttt{[MASK]} token with a probability of $1-\alpha_t$. The noising process only acts on the current block $\mathbf{b}^k$.
\paragraph{Reverse Process (Denoising Process)}
Given the historical context $\mathbf{b}^{<k}$ and the current noisy state $\mathbf{b}^k_t$, the model predicts the original state of the current block in parallel:
\begin{equation}
p_\theta(\mathbf{b}^k_0 \mid \mathbf{b}^k_t, \mathbf{b}^{<k}) = \prod_{j=1}^{B} p_\theta((\mathbf{b}^k_0)_j \mid \mathbf{b}^k_t, \mathbf{b}^{<k})
\end{equation}
As for the optimization objective, the training of dLLMs typically involves maximizing the Evidence Lower Bound (ELBO). In the blockwise diffusion framework, this translates to minimizing the Conditional Negative Evidence Lower Bound (NELBO) over all blocks, where $w(t)$ is a time-dependent weighting coefficient, and $\text{CE}$ denotes the cross-entropy loss.
\begin{equation}
\mathcal{L}_{\text{BD}}(\theta) = \sum_{k=1}^{K} \mathbb{E}_{t, \mathbf{b}^k_0} \left[ w(t) \cdot \sum_{j=1}^{B} \mathbb{1}[(\mathbf{b}^k_t)_j = \texttt{[MASK]}] \cdot \text{CE}\left( p_\theta(\cdot \mid \mathbf{b}^k_t, \mathbf{b}^{<k}), (\mathbf{b}^k_0)_j \right) \right]
\end{equation}

Thanks to the combination of intra-block AR and inner-block diffusion, the KV cache mechanism can be applied to blockwise dLLMs, thus enhancing the computational efficiency of traditional dLLMs and simplifying the likelihood computation~\citep{cheng2025sdar,wang2025diffusion}.

\subsection{RL and its application in dLLM}
Reinforcement Learning (RL) is a key training paradigm in the alignment stage of LLMs~\citep{OpenAI2024o1,guo2025deepseek,openr1}. In particular, Group Relative Policy Optimization (GRPO)~\citep{deepseekv3,guo2025deepseek} has been the RL algorithm mostly used on AR. It replaces the critic model in PPO~\citep{schulman2017proximal} by using diversified sampling and within-group normalized advantages, enabling lightweight online RL and steering LLM toward outputs that are better aligned with human preferences.
Specifically, its advantage calculation is as follows.
First, for a prompt $q$, sample $G$ outputs $\{o_1, \dots, o_G\}$ from the old policy $\pi_{\theta_{\text{old}}}$. Let the reward of completion $o_i$ be $r_i$. The group-normalized advantage is defined as $A_i = r_i - \frac{1}{G}\sum_{j=1}^{G} r_j$, and its token-level assignment is simply$A_{i,k} = A_i$, where $k = 1,\dots,|o_i|$. Since AR models provide per-token conditional probabilities, the importance sampling ratio for each token is
\begin{equation}
\rho_{i,k}
=
\frac{
\pi_{\theta}\!\left(o_{i,k} \mid q, o_{i,<k}\right)
}{
\pi_{\theta_{\text{old}}}\!\left(o_{i,k} \mid q, o_{i,<k}\right)
}.
\end{equation}
Then the GRPO objective with clipping and reverse KL regularization is defined as follows: 
\begin{equation}
\begin{aligned}
\mathcal{L}_{\mathrm{GRPO}}(\theta)
&= \mathbb{E}_{q,\,o_{1:G}} \\
&\quad \left[
\frac{1}{G}
\sum_{i=1}^{G}
\frac{1}{|o_i|}
\sum_{k=1}^{|o_i|}
\min\!\Big(
\rho_{i,k} A_{i,k},
\mathrm{clip}(\rho_{i,k},\,1-\epsilon,\,1+\epsilon) A_{i,k}
\Big)
\right]
- \beta D_{\mathrm{KL}}(\pi_{\theta} \,\|\, \pi_{\mathrm{ref}}).
\end{aligned}
\end{equation}

However, RL algorithms such as GRPO are difficult to transfer to dLLM, because dLLMs do not provide per-token probabilities directly, and their log-probability estimates require multi-step sampling, resulting in high computational cost and high variance. In addition, the random masking step map used in pre-training is also mismatched with the decoding trajectory in realistic inference. 

Thanks to the introduction of blockwise dLLMs~\citep{cheng2025sdar,wang2025diffusion}, the computational overhead of log-probability is greatly reduced. Building on this, \citet{wang2025revolutionizing} proposes TraceRL, which leverages the algorithmic efficiency of blockwise dLLMs to enable faster unbiased logit computation. However, due to the lack of attention acceleration during training and the absence of tight integration between training and inference frameworks, TraceRL still leaves significant room for engineering optimization. Moreover, it does not implement more fine-grained RL algorithms such as GRPO. To address these limitations, we introduce our efficient training framework DiRL and DiPO, the first unbiased implementation of GRPO for dLLMs.

\section{DiRL Post-Training}

Using the DiRL framework, we conduct two-stage post-training, SFT followed by RL based on our DIPO, achieve sota dLLM, DiRL-8B-Instruct.

\subsection{SFT stage}

\paragraph{SFT Data} We train on the OpenR1-Math dataset~\citep{openr1} distilled from GLM-4.6~\citep{zeng2025glm,glm46}. We chose GLM-4.6 because currently, it is almost the best-performing open-source non-reasoning LLM on math tasks and yields reasoning trajectories of manageable length. As long-reasoning evaluation for diffusion models is still scarce, we cap the reasoning length at 8k token length, which is already the longest inference length reported for dLLMs.

\paragraph{SFT setup} SFT is conducted with LLaMA-Factory~\citep{zheng2024llamafactory}. 8$\times$H200 GPUs are applied to fine-tune with SDAR-8B-Chat with DeepSpeed ZeRO1~\citep{rajbhandari2020zero}. Models are fine-tuned with a maximum length of 8k tokens. We set the global batch size to 512, the maximum learning rate to 1e-5, and the weight decay to 0, and fine-tune it in 100 steps with a cosine annealing learning rate scheduler.

\subsection{RL stage}

\paragraph{DiPO} 

Our optimization objective is defined as:
\begin{equation}
\begin{aligned}
J_{\text{policy}}(\theta_p)
&= \mathbb{E}_{Q \sim \mathcal{D}_{\text{task}}, \{\tau_i\}_{i=1}^{G} \sim \pi_{\text{old}}(\cdot \mid Q)} \\
&\quad \Bigg[
\sum_{i=1}^{G}
\frac{1}{|\tau_i|}
\sum_{t=1}^{|\tau_i|}
\sum_{o_k \in \tau_i(t)}
C_\epsilon\!\left(
\frac{
\pi_{\theta_p}(o_k \mid \tau_i(1{:}t-1))
}{
\pi_{\text{old}}(o_k \mid \tau_i(1{:}t-1))
},
A_i
\right)
- \beta \, \mathrm{KL}\!\left[\pi_{\theta_p} \,\|\, \pi_{\text{ref}}\right]
\Bigg],
\end{aligned}
\end{equation}

where we assume that each trajectory requires $t$ decoding steps, and
$\tau_i(1{:}t-1) \triangleq \bigcup_{j=1}^{\,t-1} \tau_i(j)$
denotes the prefix up to step $t-1$, namely tokens decoded by timestep $t-1$. 
To maintain stable learning, we use the clipping operator $C_\epsilon(r, A)
\triangleq
\min\big(rA,\; \mathrm{clip}(r,\,1-\epsilon,\,1+\epsilon)\,A\big)$, which limits how much the updated policy can deviate from the behavior policy at each step. The term $A_i$ stands for the normalized advantage assigned to the $i$-th trajectory, 
while $\pi_{\text{old}}$ represents the policy that produced the samples. 
In addition to the clipped surrogate, we also incorporate a KL penalty with respect to a fixed reference policy $\pi_{\text{ref}}$, rather than the behavior policy, so that the learned model does not drift too far away from the underlying reference model. 
The coefficient $\beta$ controls the strength of this regularization term.

Besides, since our framework performs online updates, the behavior policy $\pi_{\text{old}}$ is implemented as the 
detached output of the current policy $\pi_{\theta_p}$ in the optimization objective: 

\begin{equation}
\begin{aligned}
J_{\text{policy}}(\theta_p)
&= \mathbb{E}_{Q \sim \mathcal{D}_{\text{task}}, \{\tau_i\}_{i=1}^{G} \sim \pi_{\text{old}}(\cdot \mid Q)} \\
&\quad \Bigg[
\frac{1}{\sum_{i=1}^{G} |\tau_i|}
\sum_{i=1}^{G}
\sum_{t=1}^{|\tau_i|}
\sum_{o_k \in \tau_i(t)}
C_\epsilon\!\left(
\frac{
\pi_{\theta_p}(o_k \mid \tau_i(1{:}t-1))
}{
\mathrm{sg}(\pi_{\theta_p}(o_k \mid \tau_i(1{:}t-1)))
},
A_i
\right)
- \beta \, \mathrm{KL}\!\left[\pi_{\theta_p} \,\|\, \pi_{\text{ref}}\right]
\Bigg].
\end{aligned}
\end{equation}

Here $\mathrm{sg}[\cdot]$ denotes the stop-gradient operator (equivalently \texttt{.detach()} in PyTorch). 

Moreover, we integrate the DAPO\citep{yu2025dapoopensourcellmreinforcement} algorithm, which uses the token-level policy gradient:

\begin{equation}
\begin{aligned}
J_{\text{policy}}(\theta_p) 
&= \mathbb{E}_{Q \sim \mathcal{D}_{\text{task}}, \{\tau_i\}_{i=1}^{G} \sim \pi_{\text{old}}(\cdot \mid Q)} \\
&\quad \Bigg[
  \frac{1}{\sum_{i=1}^{G} |\tau_i|}
  \sum_{i=1}^{G} \sum_{t=1}^{|\tau_i|}
  C_\epsilon\!\left(
    \frac{
      \pi_{\theta_p}\!\big(\tau_i(t)\mid \tau_i(1{:}t-1)\big)
    }{
      sg(\pi_{\theta_p}\!\big(\tau_i(t)\mid \tau_i(1{:}t-1)\big))
    },
    A_i
  \right)
\Bigg].
\end{aligned}
\end{equation}

\paragraph{RL Data} We train on the Big-Math dataset~\citep{albalak2025big}, a large-scale, high-quality math dataset verfied by math-verify, and specially designed for RL.

\paragraph{RL setup} RL is conducted after SFT with \textbf{DiRL} supported by Accelerate. 128$\times$H200 GPUs are applied to train the model with DeepSpeed ZeRO1. Models are fine-tuned with a maximum length of 8k tokens. We set the global batch size to 128, and rollout 32 trajectories for each problem. The learning rate is set to 1e-6, and the weight decay to 0, and train it in 40 steps.

\section{Engineering Optimization}

\begin{figure}[!t]
\begin{minipage}{0.98\textwidth}
    \hspace*{0.8cm}
    \centering
    \begin{subfigure}[b]{0.38\linewidth}
        \centering
        \includegraphics[width=\linewidth]{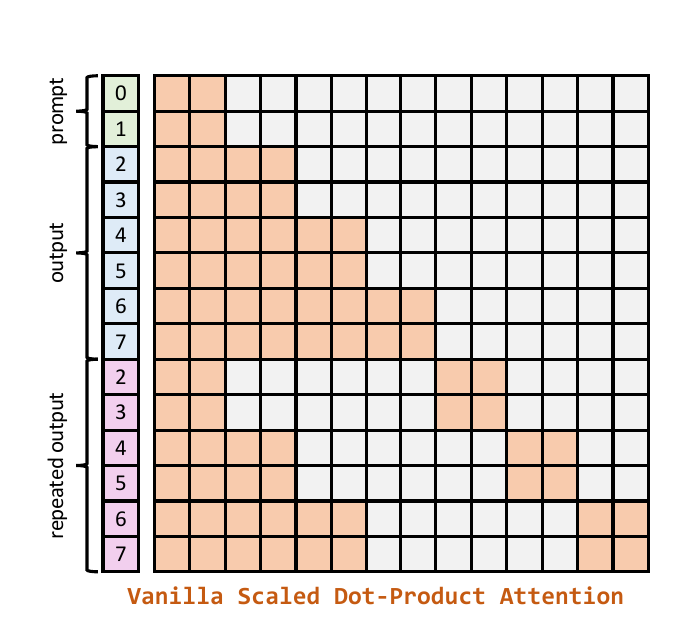}
        \caption{Attention mask in TraceRL.}
        \label{fig_attn_mask_current}
    \end{subfigure}
    \hfill
    \begin{subfigure}[b]{0.38\linewidth}
        \centering
        \includegraphics[width=\linewidth]{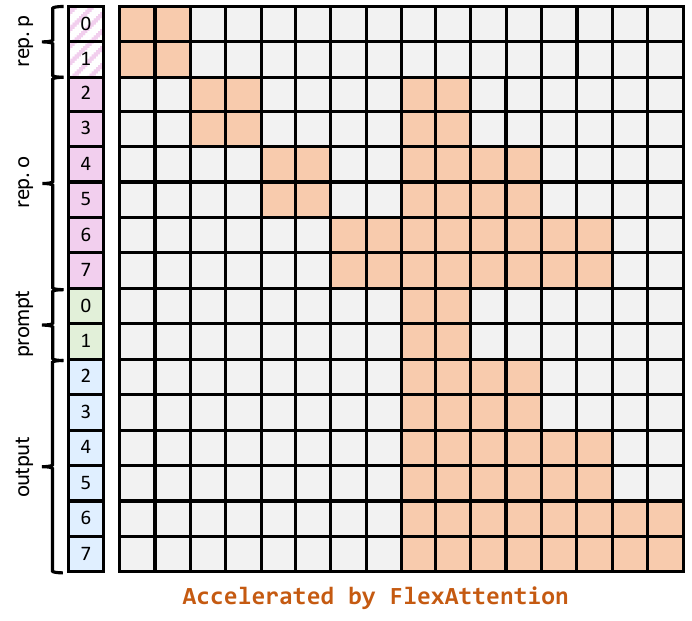}
        \caption{Attention mask in our DiRL.}
        \label{fig_attn_mask_dirl}
    \end{subfigure}
    \hspace*{0.8cm}
    \caption{The visualized comparison of the attention mask between our DiRL framework and TraceRL, where block size is 2, the length of prompt colored in green is 2, and the length of output colored in blue is 6. The loss is calculated based on the repeated output part colored in full purple.\label{fig_attn_mask}}
\end{minipage}
\end{figure}

As shown in Figure~\ref{dirl_framework}, the DiRL framework incorporates efficiency designs at both the model and system levels. At the model level, blockwise dLLM~\citep{cheng2025sdar} with FlexAttention~\citep{dong2024flex} compute unbiased logits efficiently, and at the framework level, tight integration of the LMDeploy inference engine~\citep{2023lmdeploy} with the open-source distributed training library LLaMA-Factory~\citep{zheng2024llamafactory} enables online model updates for highly efficient training.

\subsection{Blockwise dLLM with FlexAttention}

At the model level, as mentioned earlier, we use blockwise dLLM, which is more efficient for logic computation than the original dLLMs in post-training. This is because it maintains an AR paradigm between blocks, performing denoising only within each block. Thanks to this design, as Figure~\ref{fig_attn_mask_current} shows, TraceRL~\citep{wang2025revolutionizing} can parallelize denoising across blocks by repeating the output twice and altering the attention mask, thus enabling faster SFT training.

However, the dominant FlashAttention interface~\citep{dao2022flashattention,dao2023flashattention} cannot handle the complex attention masks required by blocksize dLLMs, limiting the practical training throughput. Fortunately, PyTorch’s FlexAttention~\citep{dong2024flex} fills this gap with an efficient operator that accepts fine-grained masks. DiRL does not differentiate between the prompt and output, repeats both parts blockwise, and reshapes the SFT mask, as illustrated in Figure~\ref{fig_attn_mask_dirl}. Based on a more regular attention mask, we combines it with FlexAttention, cutting post-training latency dramatically.

\subsection{Training and Inference Integration}

\begin{figure}[!t]
\begin{minipage}{0.98\textwidth}
    \hspace*{0.4cm}
    \begin{subfigure}[b]{0.45\linewidth}
        \centering
        \includegraphics[width=\linewidth]{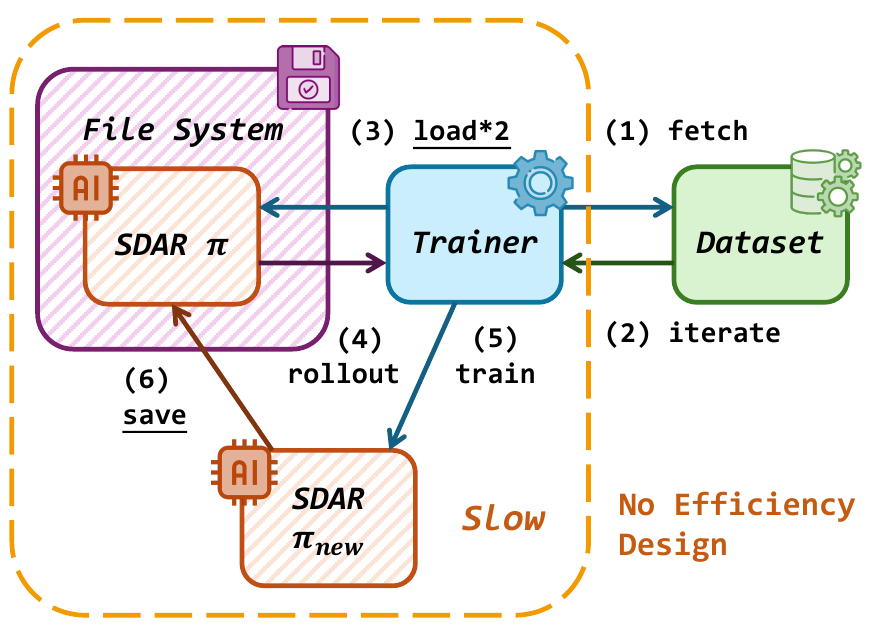}
        \caption{RL in current dLLMs.}
        \label{fig_dllm_rl_current}
    \end{subfigure}
    \hfill
    \begin{subfigure}[b]{0.45\linewidth}
        \centering
        \includegraphics[width=\linewidth]{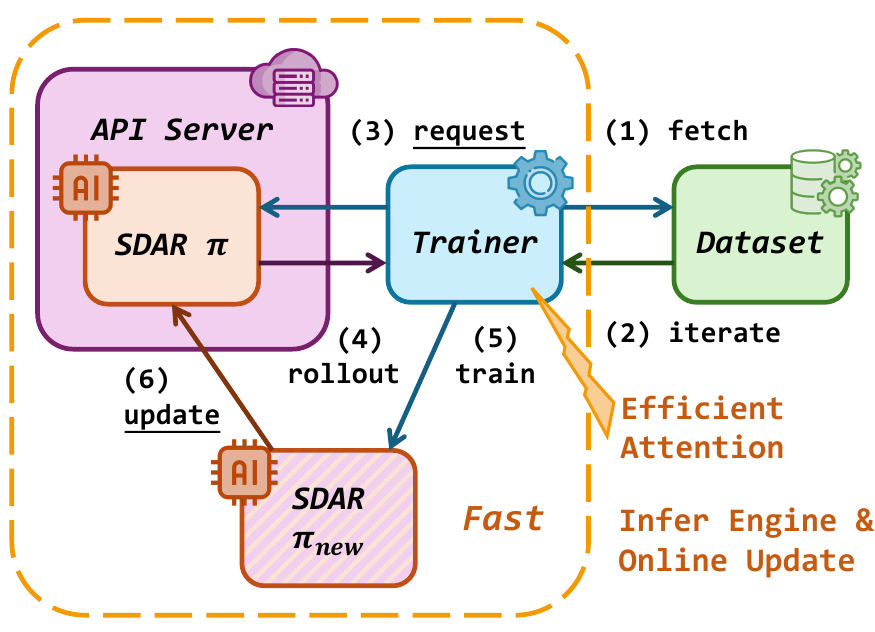}
        \caption{RL in our DiRL Framework.}
        \label{fig_dllm_rl_dirl}
    \end{subfigure}
    \hspace*{0.4cm}
    \caption{The training and inference integration of our DiRL compared to RL in current dLLMs.\label{fig_dllm_rl}}
\end{minipage}
\end{figure}

Existing RL efforts in dLLMs~\citep{wang2025revolutionizing}, as noted earlier, lack inference-engine support and co-design optimization. Concretely, Figure~\ref{fig_dllm_rl_current} illustrates the online-learning loop in existing dLLM-oriented RL. Each training step fetches new data, loads the previous checkpoint for inference, conducts rollout, loads the checkpoint twice for training, trains, and then saves the updated checkpoint back to the file system for reloading in the next step, resulting in an evident waste of IO interactions. Moreover, in earlier works~\citep{zhao2025d1,zhu2025llada}, without the support of an inference engine, the inference of dLLM is slow, further amplifying the overall training inefficiency.

\begin{figure}[!t]
    \begin{minipage}[b]{0.65\linewidth}
        \centering
        \includegraphics[width=\linewidth]{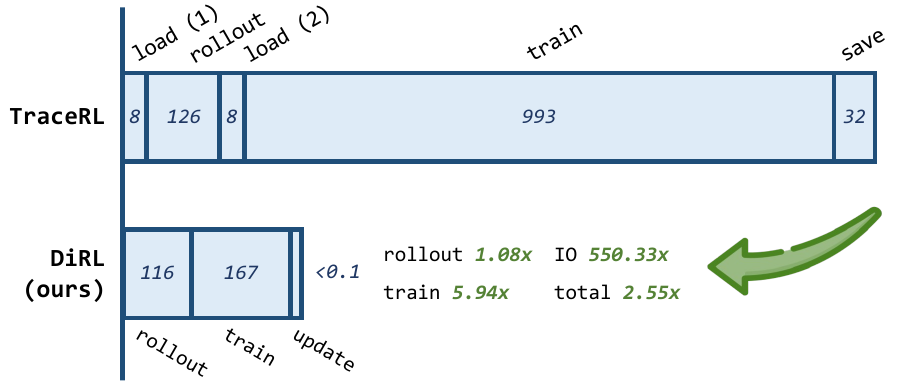}
        \caption{Comparison of time breakdown per RL training step for SDAR-8B-Chat, measured in seconds. The comparison is conducted on 8$\times$H200 GPUs with batch size 4, input length 1024, and output length 8192.}
        \label{fig_breakdown}
    \end{minipage}
    \hfill
    \begin{minipage}[b]{0.32\linewidth}
        \centering
        \includegraphics[width=\linewidth]{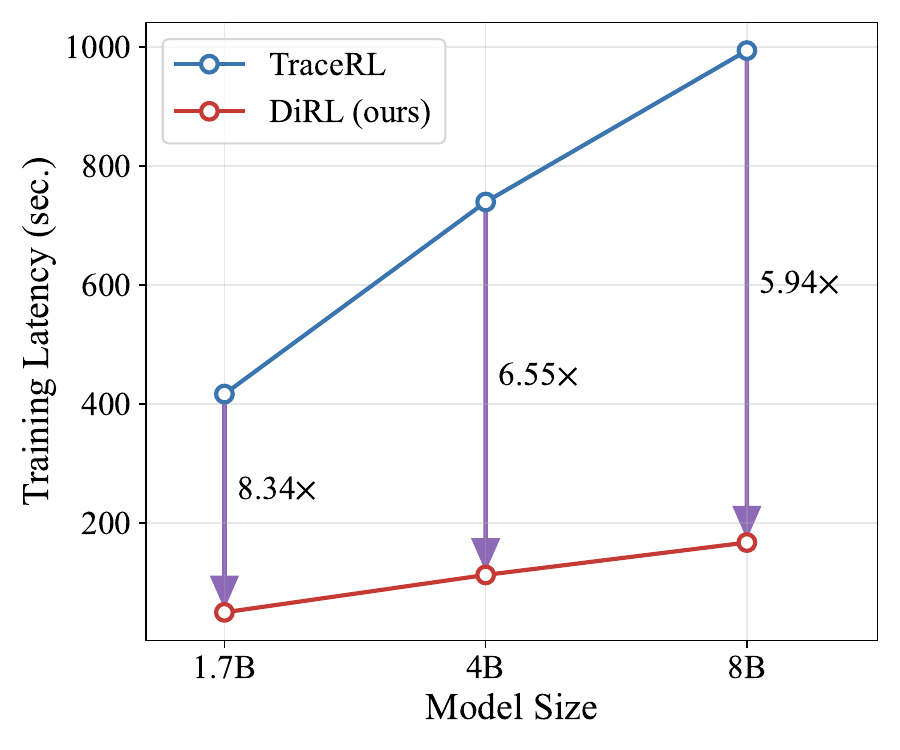}
        \caption{Reduction of training latency of our DiRL across different sizes of SDAR models.}
        \label{dirl_train_tgs}
    \end{minipage}
\end{figure}

Compared with existing approaches, DiRL tightly couples training and inference as shown in Figure~\ref{fig_dllm_rl_dirl}. We first deploy the model in an API server through LMDeploy~\citep{2023lmdeploy}, which is the only loading operation in the whole training process. Leveraging LMDeploy’s in-place parameter-update API, we immediately push each training-step checkpoint into the server, eliminating IO between the file system while keeping the API server alive. As shown in Figure~\ref{fig_breakdown}, these refinements cut per-step latency sharply and dramatically boost the efficiency of dLLM-oriented RL.

\subsection{Efficiency Validation}

We compare the per-step time breakdown of DiRL and TraceRL on 8$\times$H200 GPUs with batch size 4, input length 1024 and output length 8192 in Figure~\ref{fig_breakdown}. Since TraceRL also uses the JetEngine rollout backend~\citep{cheng2025sdar}, the speed-up in that operation is the modest. However, FlexAttention-accelerated training reduces latency by nearly 6$\times$. Furthermore, Figure~\ref{dirl_train_tgs} shows that this gain persists across model sizes. 8B training latency of our DiRL per step is lower than 1.7B latency of TraceRL. Moreover, replacing two model loads and one save with an almost cost-free in-place update yields an overall 2.5$\times$ throughput improvement.

\subsection{Main Results}

We evaluate our DiRL-8B-Instruct with other baselines on five representative math tasks, MATH500~\citep{hendrycks2021math}, GSM8k~\citep{cobbe2021gsm8k}, AIME2024~\citep{aime2024}, AIME2025~\citep{aime2025}, OlympiadBench~\citep{he2024olympiadbench}. Baselines include the blockwise dLLMs SDAR-8B-Chat~\citep{cheng2025sdar} and TraDo-8B-Instruct~\citep{wang2025revolutionizing}. We report results for both static and dynamic decoding. For the latter, we set a threshold of 0.9, decoding tokens whose top-1 probability exceeds 0.9 directly. As an external reference, we also list the widely used autoregressive LLM Qwen2.5 Series~\citep{qwen2024qwen25technicalreport}. The results are shown in Table~\ref{tab_main}.

Our DiRL-8B-Instruct achieves the best result on the average score and the majority of tasks, outperforming blockwise dLLMs and AR models with the same model size, even exceeding the larger Qwen2.5-32B-Instruct, and establishing a new state-of-the-art dLLM. Comparing average reasoning lengths across tasks, DiRL-8B-Instruct produces the longest output, indicating that the two-stage SFT+RL post-training equips it with stronger math reasoning capability and enables more complex math derivations that yield superior performance.

\begin{table*}[!t]
\tabcolsep=0.07cm
\centering
\small
\begin{tabular}{l ccc ccc ccc ccc ccc c}
\toprule
& \multicolumn{3}{c}{\textbf{MATH500}} & \multicolumn{3}{c}{\textbf{GSM8k}} & \multicolumn{3}{c}{\textbf{AIME2024}} & \multicolumn{3}{c}{\textbf{AIME2025}} & \multicolumn{3}{c}{\textbf{Olympiad}} & \textbf{Avg.} \\ 
\midrule
\textbf{\textit{Qwen2.5-7B-Instruct}} & 73.8 & $_\text{1.0}$ & $_\text{628.1}$ & 89.8 & $_\text{1.0}$ & $_\text{308.9}$ & 9.0 & $_\text{1.0}$ & $_\text{1005.5}$ & 5.6 & $_\text{1.0}$ & $_\text{970.5}$ & 36.6 & $_\text{1.0}$ & $_\text{862.0}$ & 42.9 \\
\textbf{\textit{Qwen2.5-32B-Instruct}} & 81.1 & $_\text{1.0}$ & $_\text{554.6}$ & \textbf{94.0} & $_\text{1.0}$ & $_\text{291.2}$ & 12.9 & $_\text{1.0}$ & $_\text{831.7}$ & 11.9 & $_\text{1.0}$ & $_\text{839.7}$ & 45.7 & $_\text{1.0}$ & $_\text{742.4}$ & 49.1 \\
\midrule
\textbf{\textit{SDAR-8B-Chat}} & 71.5 & $_\text{1.0}$ & $_\text{603.4}$ & 89.5 & $_\text{1.0}$ & $_\text{712.9}$ & 5.6 & $_\text{1.0}$ & $_\text{1342.4}$ & 8.5 & $_\text{1.0}$ & $_\text{920.8}$ & 35.6 & $_\text{1.0}$ & $_\text{890.4}$ & 42.2 \\
+ Dynamic & 71.9 & $_\text{2.4}$ & $_\text{616.6}$ & 89.9 & $_\text{2.8}$ & $_\text{708.8}$ & 9.2 & $_\text{2.4}$ & $_\text{1274.2}$ & 9.4 & $_\text{2.1}$ & $_\text{889.8}$ & 36.0 & $_\text{2.4}$ & $_\text{896.4}$ & 43.3 \\
\textbf{\textit{TraDo-8B-Instruct}} & 76.7 & $_\text{1.0}$ & $_\text{618.1}$ & 90.4 & $_\text{1.0}$ & $_\text{324.7}$ & 11.5 & $_\text{1.0}$ & $_\text{1036.1}$ & 13.5 & $_\text{1.0}$ & $_\text{988.0}$ & 40.2 & $_\text{1.0}$ & $_\text{864.3}$ & 46.5 \\
+ Dynamic & 75.6 & $_\text{2.3}$ & $_\text{618.6}$ & 91.1 & $_\text{2.1}$ & $_\text{315.2}$ & 11.7 & $_\text{2.0}$ & $_\text{1091.4}$ & 15.0 & $_\text{1.9}$ & $_\text{996.2}$ & 40.3 & $_\text{2.1}$ & $_\text{868.0}$ & 46.7 \\
\textbf{\textit{DiRL-8B-Instruct}} & \textbf{85.1} & $_\text{1.0}$ & $_\text{1917.2}$ & \underline{93.1} & $_\text{1.0}$ & $_\text{707.4}$ & \textbf{21.5} & $_\text{1.0}$ & $_\text{5434.2}$ & \textbf{22.9} & $_\text{1.0}$ & $_\text{5019.0}$ & \textbf{47.3} & $_\text{1.0}$ & $_\text{3556.4}$ & \textbf{54.0} \\
+ Dynamic & \underline{83.1} & $_\text{2.0}$ & $_\text{2000.7}$ & 93.0 & $_\text{2.2}$ & $_\text{730.1}$ & \underline{20.6} & $_\text{1.7}$ & $_\text{5468.5}$ & \underline{20.8} & $_\text{1.7}$ & $_\text{5129.6}$ & \underline{46.4} & $_\text{1.8}$ & $_\text{3614.5}$ & \underline{52.8} \\
\bottomrule
\end{tabular}
\caption{Comprehensive benchmark results of our DiRL-8B-Instruct compared with current dLLMs, SDAR-8B-Chat~\citep{cheng2025sdar} and TraDo-8B-Instruct~\citep{wang2025revolutionizing}, as well as well-acknowledged AR model Qwen2.5 Series~\citep{qwen2024qwen25technicalreport}. Each cell presents the accuracy, with best values in bold and suboptimal values underlined, as well as the average number of decoding tokens per step and the average output lengths.}
\label{tab_main}
\end{table*}

\section{Discussion}

\subsection{Ablation Study}

\begin{figure}[!t]
    \centering
    
    \begin{subfigure}[b]{0.325\textwidth}
        \centering
        \includegraphics[width=\textwidth]{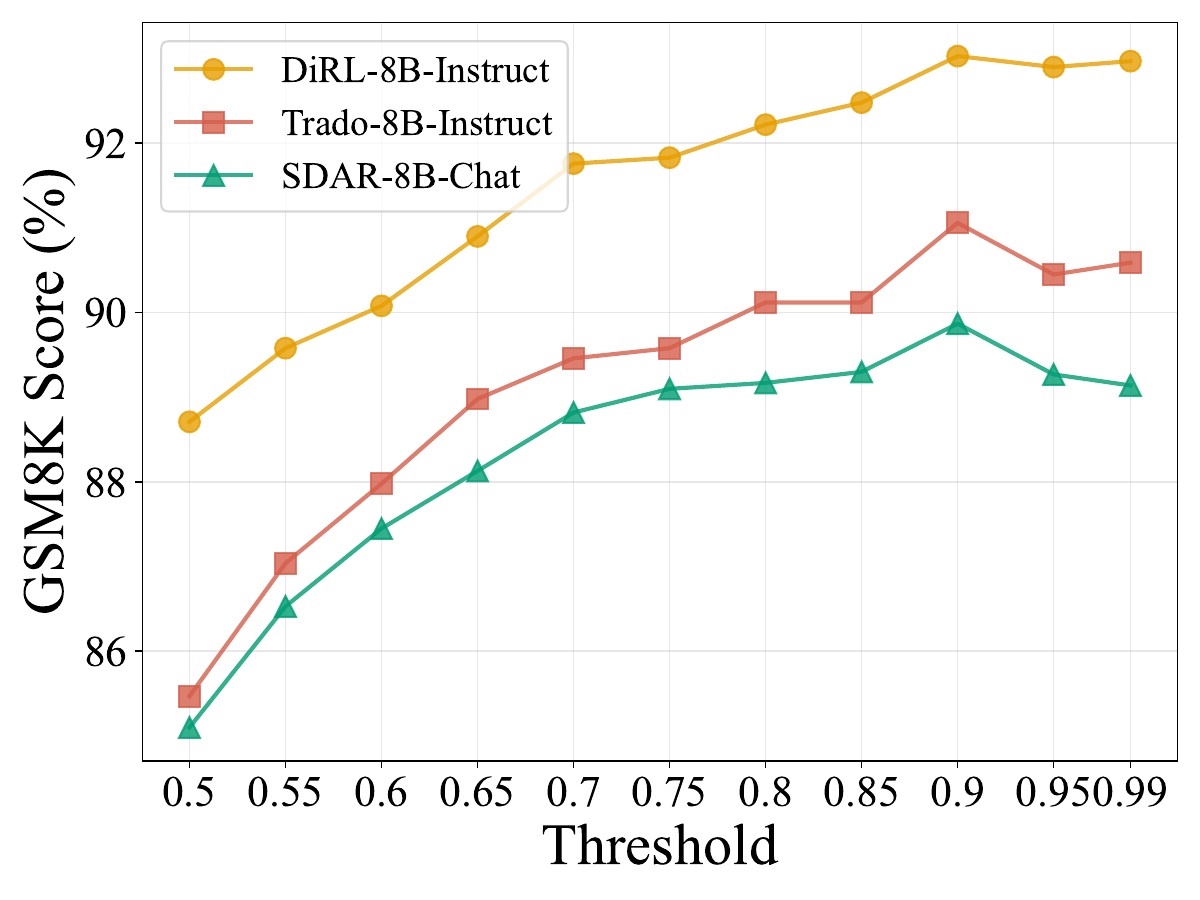}
        \caption{GSM8K}
        \label{GSM8K}
    \end{subfigure}
    \hfill
    \begin{subfigure}[b]{0.325\textwidth}
        \centering
        \includegraphics[width=\textwidth]{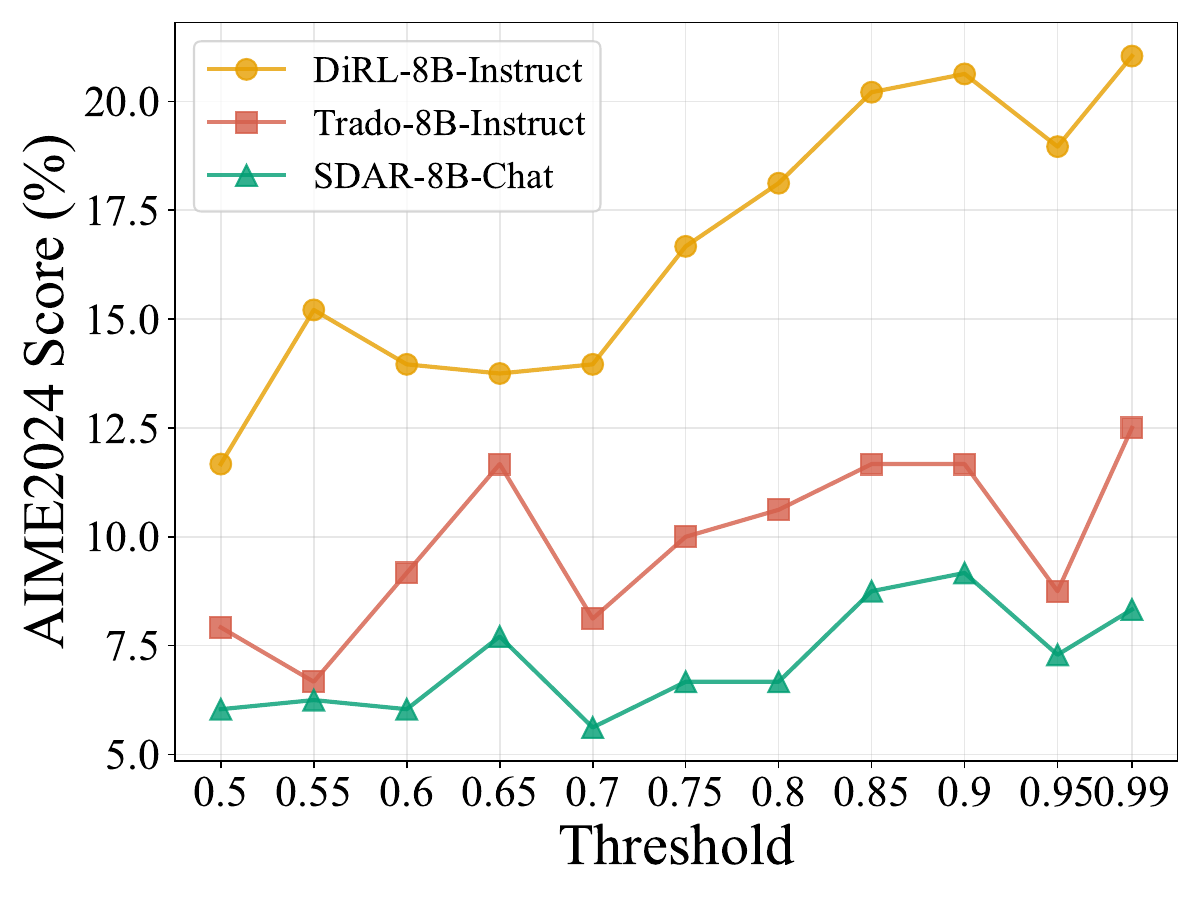}
        \caption{AIME2024}
        \label{AIME2024}
    \end{subfigure}
    \hfill
    \begin{subfigure}[b]{0.325\textwidth}
        \centering
        \includegraphics[width=\textwidth]{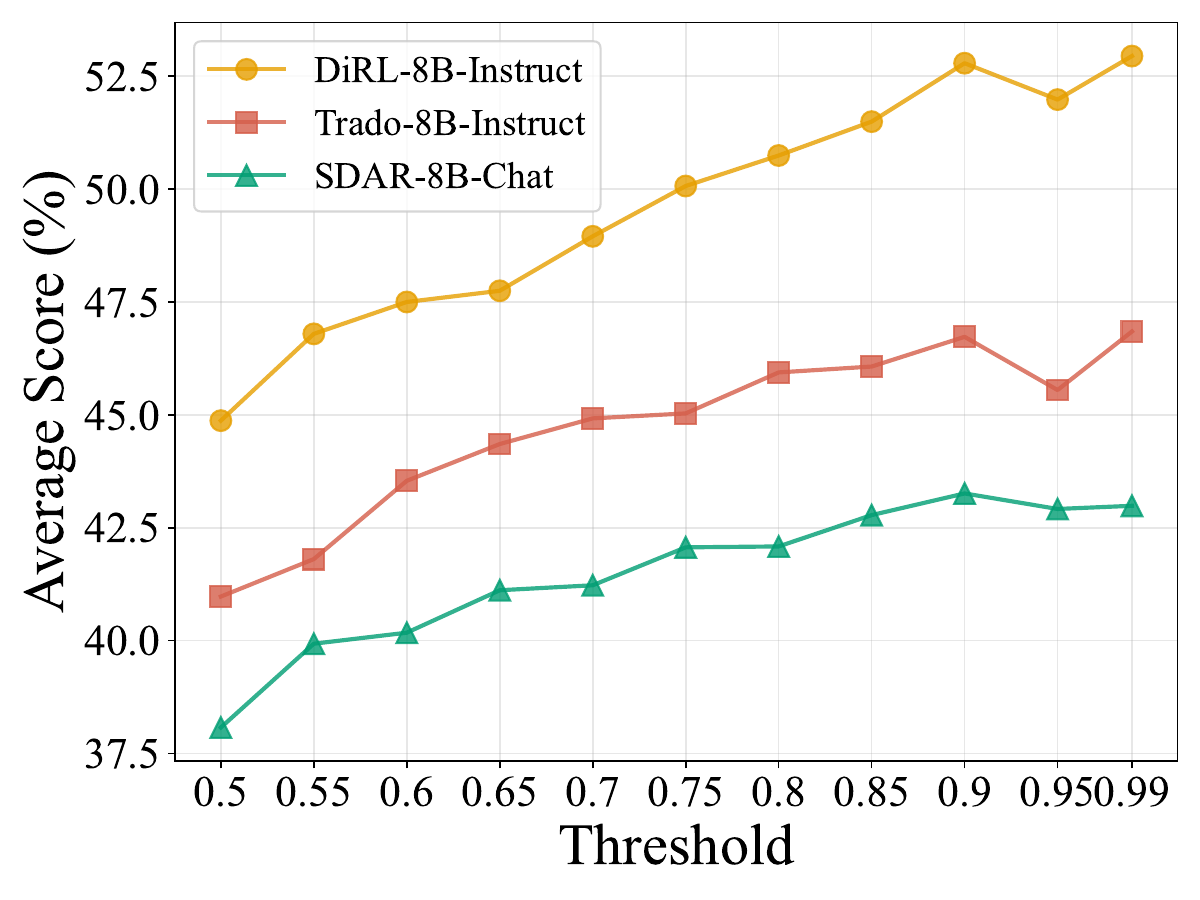}
        \caption{Average}
    \end{subfigure}

    \caption{Ablation study on dynamic sampling threshold}
    \label{ablation}
\end{figure}

To investigate the sensitivity of the models to hyperparameters within the dynamic decoding strategy, we perform an ablation study on the dynamic sampling threshold $\tau$, adjusting it within the range of $0.5$ to $0.99$, and record the accuracy and average scores of DiRL-8B-Instruct, TraDo-8B-Instruct~\citep{wang2025revolutionizing}, and SDAR-8B-Chat~\citep{cheng2025sdar} on five mathematical benchmarks (as illustrated in Figure~\ref{ablation}). 

Experimental results indicate that regardless of the variation in $\tau$, DiRL-8B-Instruct consistently and significantly outperforms the baseline models in all individual tasks and average performance, demonstrating strong robustness under varying degrees of aggressive decoding. Specifically, as the threshold increases (implying a more conservative decoding approach closer to greedy search), the performance of DiRL-8B-Instruct exhibits a steady upward trend. In contrast, the responses of other dLLMs are suboptimal. SDAR-8B-Chat displays performance plateu across most tasks, suggesting that mere threshold adjustment cannot compensate for its inherent reasoning deficiencies, while TraDo-8B-Instruct, despite acceptable performance on GSM8K~\ref{GSM8K}, suffers from significant fluctuations in tasks such as AIME2024~\ref{AIME2024}, indicating a lack of stability. 

These findings not only confirm that the superior mathematical reasoning capacity of DiRL-8B-Instruct stems from the high-quality reasoning paths derived from the two-stage SFT+RL post-training rather than the dependency on specific hyperparameter configurations, but also reveal that $\tau=0.9$ serves as the optimal equilibrium point for average performance across models, thus further validating the rationality of parameter selection in our main experiments.

\subsection{Future work}

Despite the clear gain in training efficiency and the current best-result at 8B scale, several improvements remain for our future work. First, we will scale the approach to larger dLLMs to pursue even stronger performance. Second, although our 8k inference length is already the longest reported for dLLMs, it is still short compared with AR models~\citep{guo2025deepseek,OpenAI2024o1}. We have not yet explored test-time scaling or long CoT techniques that could raise the ceiling of model itself~\citep{liu2025thus,zeng2024scaling,zeng2025revisiting}. Third, we will incorporate strategies such as dynamic packing~\citep{bai2024longalign} from AR long-context training to further accelerate our framework. Finally, though currently focused on math tasks, we will extend evaluation to agentic and code tasks~\citep{zeng2025glm}, striving to match or surpass AR models across a broader spectrum.

\section{Conclusion}

In this work, we introduce DiRL, an efficient framework for SFT and RL in dLLMs. By combining FlexAttention with refined attention masks, we remarkably reduce the training latency of blockwise dLLMs. By applying LMDeploy and online parameter updates, we achieve an unbiased logits computation and efficient two-stage post-training. Building upon this foundation, we propose DiPO, the first unbiased GRPO implementation for dLLMs, and train DiRL-8B-Instruct, the currently strongest dLLM on math tasks, even outperforming the Qwen2.5 series. We will continue to refine our framework for higher training efficiency and validate our approach on larger models, longer reasoning chains, and a broader downstream tasks, aiming to provide the community with a solid baseline and an open-source post-training toolkit for dLLMs.

\bibliography{dirl}
\bibliographystyle{dirl}
\appendix

\section{More Related Work of RL in dLLMs}

Besides TraceRL~\citep{wang2025revolutionizingreinforcementlearningframework} targeted on blockwise dLLMs, more works have made contributions to adapting RL to Diffusion LLMs.
diffu-GRPO~\citep{zhao2025d1scalingreasoningdiffusion} uses a single forward pass to perform “one-step decoding” on masked completion tokens (with optional random prompt masking), approximating the sequence log-probability as the sum of per-token log-probabilities, which enables low-cost policy optimization but introduces substantial approximation bias.
To mitigate this issue, Coupled-GRPO~\citep{gong2025diffucoderunderstandingimprovingmasked} constructs a pair of negatively correlated estimation samples by applying complementary masks at (t, T-t), thereby reducing the Monte Carlo variance of token log-probability estimates and stabilizing GRPO updates.

Similarly, when performing DPO alignment for diffusion LLMs using ELBO to approximate log-likelihood---which introduces high variance and unstable training---VRPO~\citep{zhu2025vrporethinkingvaluemodeling} systematically reduces the variance of the preference score by increasing the ELBO sampling budget, allocating the entire budget to different timesteps ($n_t = n$, $n_{y_t} = 1$), and letting the current and reference model share the same batch of ($t$, masked $y_t$) samples for antithetic sampling.
However, because VRPO~\citep{zhu2025vrporethinkingvaluemodeling} must keep the computation graphs of all Monte Carlo samples in memory at once, it easily leads to GPU memory blow-ups and limits the number of samples that can be used. 
To resolve this, BGPO~\citep{lin2025boundaryguidedpolicyoptimizationmemoryefficient} reformulates VRPO’s objective—“taking the exponential of the average log-likelihood difference over all time samples,” which couples all samples—into a linearly separable lower bound over individual samples, enabling per-sample backpropagation and gradient accumulation and reducing memory usage from $O(n_t)$ to $O(1)$. Other variance-reduction methods optimize sampling strategies, such as GDPO~\citep{rojas2025improvingreasoningdiffusionlanguage}, which fixes the noise levels $t$ to a small set of Gauss–quadrature points and performs one-step parallel denoising on generated answers to approximate the sequence-level ELBO, thereby enabling low-variance sequence-level importance ratios.

However, in the above approaches, random sampling and random masking do not seem to resolve the mismatch between the reinforcement learning objective and the actual reasoning trajectory. TraceRL~\citep{wang2025revolutionizingreinforcementlearningframework} abandons the “random mask + one-step denoise” scheme for obtaining importance sampling ratios. Instead, it segments the generated sequence into trajectories by step/block, uses a shrinkage parameter s to aggregate neighboring steps, and computes token-level ratios within each step by weighting with the old/new policy probability ratios. It then propagates learning signals step by step along the generation process, while introducing a diffusion-based value model to construct step-level GAE advantages.

In addition, to reduce the ``zero-gradient'' issue in GRPO, IGPO~\citep{zhao2025inpaintingguidedpolicyoptimizationdiffusion} uses ground-truth fragments as controllable hints to perform inpainting on fully incorrect samples, generating learnable correct trajectories and stabilizing their influence through entropy filtering, thereby restoring effective gradients for diffusion LLMs under sparse rewards.

\end{document}

%% file: math_commands.tex
\usepackage{amsmath,amsfonts,bm}

\def\eqref#1{equation~\ref{#1}}

\def\1{\bm{1}}

\DeclareMathAlphabet{\mathsfit}{\encodingdefault}{\sfdefault}{m}{sl}
\SetMathAlphabet{\mathsfit}{bold}{\encodingdefault}{\sfdefault}{bx}{n}

%% file: dirl.bib
@article{liu2025thus,
  title={Thus spake long-context large language model},
  author={Liu, Xiaoran and Li, Ruixiao and Huang, Mianqiu and Liu, Zhigeng and Song, Yuerong and Guo, Qipeng and He, Siyang and Wang, Qiqi and Li, Linlin and Liu, Qun and others},
  journal={arXiv preprint arXiv:2502.17129},
  year={2025}
}

@article{zeng2024scaling,
  title={Scaling of search and learning: A roadmap to reproduce o1 from reinforcement learning perspective},
  author={Zeng, Zhiyuan and Cheng, Qinyuan and Yin, Zhangyue and Wang, Bo and Li, Shimin and Zhou, Yunhua and Guo, Qipeng and Huang, Xuanjing and Qiu, Xipeng},
  journal={arXiv preprint arXiv:2412.14135},
  year={2024}
}

@article{zeng2025revisiting,
  title={Revisiting the Test-Time Scaling of o1-like Models: Do they Truly Possess Test-Time Scaling Capabilities?},
  author={Zeng, Zhiyuan and Cheng, Qinyuan and Yin, Zhangyue and Zhou, Yunhua and Qiu, Xipeng},
  journal={arXiv preprint arXiv:2502.12215},
  year={2025}
}

@article{bai2024longalign,
  title={Longalign: A recipe for long context alignment of large language models},
  author={Bai, Yushi and Lv, Xin and Zhang, Jiajie and He, Yuze and Qi, Ji and Hou, Lei and Tang, Jie and Dong, Yuxiao and Li, Juanzi},
  journal={arXiv preprint arXiv:2401.18058},
  year={2024}
}

@article{liu2025longllada,
  title={Longllada: Unlocking long context capabilities in diffusion llms},
  author={Liu, Xiaoran and Song, Yuerong and Liu, Zhigeng and Huang, Zengfeng and Guo, Qipeng and He, Ziwei and Qiu, Xipeng},
  journal={arXiv preprint arXiv:2506.14429},
  year={2025}
}

@article{song2025sparse,
  title={Sparse-dllm: Accelerating diffusion llms with dynamic cache eviction},
  author={Song, Yuerong and Liu, Xiaoran and Li, Ruixiao and Liu, Zhigeng and Huang, Zengfeng and Guo, Qipeng and He, Ziwei and Qiu, Xipeng},
  journal={arXiv preprint arXiv:2508.02558},
  year={2025}
}

@article{nie2025large,
  title={Large language diffusion models},
  author={Nie, Shen and Zhu, Fengqi and You, Zebin and Zhang, Xiaolu and Ou, Jingyang and Hu, Jun and Zhou, Jun and Lin, Yankai and Wen, Ji-Rong and Li, Chongxuan},
  journal={arXiv preprint arXiv:2502.09992},
  year={2025}
}

@article{zhu2025llada,
  title={LLaDA 1.5: Variance-Reduced Preference Optimization for Large Language Diffusion Models},
  author={Zhu, Fengqi and Wang, Rongzhen and Nie, Shen and Zhang, Xiaolu and Wu, Chunwei and Hu, Jun and Zhou, Jun and Chen, Jianfei and Lin, Yankai and Wen, Ji-Rong and others},
  journal={arXiv preprint arXiv:2505.19223},
  year={2025}
}

@article{ye2025dream,
  title={Dream 7b: Diffusion large language models},
  author={Ye, Jiacheng and Xie, Zhihui and Zheng, Lin and Gao, Jiahui and Wu, Zirui and Jiang, Xin and Li, Zhenguo and Kong, Lingpeng},
  journal={arXiv preprint arXiv:2508.15487},
  year={2025}
}

@misc{mercury,
    title = {Introducing Mercury, the world’s first commercial-scale diffusion language model},
    url = {https://www.inceptionlabs.ai/introducing-mercury},
    author = {Inception},
    year = {2025}
}

@misc{genimid,
    title = {Gemini Diffusion, our state-of-the-art, experimental text diffusion model},
    url = {https://deepmind.google/models/gemini-diffusion/},
    author = {Gemini},
    year = {2025}
}

@article{wang2025revolutionizing,
  title={Revolutionizing reinforcement learning framework for diffusion large language models},
  author={Wang, Yinjie and Yang, Ling and Li, Bowen and Tian, Ye and Shen, Ke and Wang, Mengdi},
  journal={arXiv preprint arXiv:2509.06949},
  year={2025}
}

@article{cheng2025sdar,
  title={SDAR: A Synergistic Diffusion-AutoRegression Paradigm for Scalable Sequence Generation},
  author={Cheng, Shuang and Bian, Yihan and Liu, Dawei and Zhang, Linfeng and Yao, Qian and Tian, Zhongbo and Wang, Wenhai and Guo, Qipeng and Chen, Kai and Qi, Biqing and others},
  journal={arXiv preprint arXiv:2510.06303},
  year={2025}
}

@article{arriola2025block,
  title={Block diffusion: Interpolating between autoregressive and diffusion language models},
  author={Arriola, Marianne and Gokaslan, Aaron and Chiu, Justin T and Yang, Zhihan and Qi, Zhixuan and Han, Jiaqi and Sahoo, Subham Sekhar and Kuleshov, Volodymyr},
  journal={arXiv preprint arXiv:2503.09573},
  year={2025}
}

@article{wang2025diffusion,
  title={Diffusion llms can do faster-than-ar inference via discrete diffusion forcing},
  author={Wang, Xu and Xu, Chenkai and Jin, Yijie and Jin, Jiachun and Zhang, Hao and Deng, Zhijie},
  journal={arXiv preprint arXiv:2508.09192},
  year={2025}
}

@article{albalak2025big,
  title={Big-math: A large-scale, high-quality math dataset for reinforcement learning in language models},
  author={Albalak, Alon and Phung, Duy and Lile, Nathan and Rafailov, Rafael and Gandhi, Kanishk and Castricato, Louis and Singh, Anikait and Blagden, Chase and Xiang, Violet and Mahan, Dakota and others},
  journal={arXiv preprint arXiv:2502.17387},
  year={2025}
}

@misc{openr1,
  title={OpenR1 Math 220k},
  author={OpenR1}, 
  year={2025}, 
  url={https://huggingface.co/datasets/open-r1/OpenR1-Math-220k}
}

@article{you2025llada,
  title={LLaDA-V: Large Language Diffusion Models with Visual Instruction Tuning},
  author={You, Zebin and Nie, Shen and Zhang, Xiaolu and Hu, Jun and Zhou, Jun and Lu, Zhiwu and Wen, Ji-Rong and Li, Chongxuan},
  journal={arXiv preprint arXiv:2505.16933},
  year={2025}
}

@article{yang2025mmada,
  title={Mmada: Multimodal large diffusion language models},
  author={Yang, Ling and Tian, Ye and Li, Bowen and Zhang, Xinchen and Shen, Ke and Tong, Yunhai and Wang, Mengdi},
  journal={arXiv preprint arXiv:2505.15809},
  year={2025}
}

@article{zhao2025d1,
  title={d1: Scaling reasoning in diffusion large language models via reinforcement learning},
  author={Zhao, Siyan and Gupta, Devaansh and Zheng, Qinqing and Grover, Aditya},
  journal={arXiv preprint arXiv:2504.12216},
  year={2025}
}

@article{ma2025dkv,
  title={dkv-cache: The cache for diffusion language models},
  author={Ma, Xinyin and Yu, Runpeng and Fang, Gongfan and Wang, Xinchao},
  journal={arXiv preprint arXiv:2505.15781},
  year={2025}
}

@article{liudllm,
  title={dLLM-Cache: Accelerating Diffusion Large Language Models with Adaptive Caching},
  author={Liu, Zhiyuan and Yang, Yicun and Zhang, Yaojie and Chen, Junjie and Zou, Chang and Wei, Qingyan and Wang, Shaobo and Zhang, Linfeng}
}

@article{wu2025fast,
  title={Fast-dLLM: Training-free Acceleration of Diffusion LLM by Enabling KV Cache and Parallel Decoding},
  author={Wu, Chengyue and Zhang, Hao and Xue, Shuchen and Liu, Zhijian and Diao, Shizhe and Zhu, Ligeng and Luo, Ping and Han, Song and Xie, Enze},
  journal={arXiv preprint arXiv:2505.22618},
  year={2025}
}

@article{wu2025fast2,
  title={Fast-dllm v2: Efficient block-diffusion llm},
  author={Wu, Chengyue and Zhang, Hao and Xue, Shuchen and Diao, Shizhe and Fu, Yonggan and Liu, Zhijian and Molchanov, Pavlo and Luo, Ping and Han, Song and Xie, Enze},
  journal={arXiv preprint arXiv:2509.26328},
  year={2025}
}

@article{he2025ultrallada,
  title={Ultrallada: Scaling the context length to 128k for diffusion large language models},
  author={He, Guangxin and Nie, Shen and Zhu, Fengqi and Zhao, Yuankang and Bai, Tianyi and Yan, Ran and Fu, Jie and Li, Chongxuan and Yuan, Binhang},
  journal={arXiv preprint arXiv:2510.10481},
  year={2025}
}

@article{nie2024scaling,
  title={Scaling up Masked Diffusion Models on Text},
  author={Nie, Shen and Zhu, Fengqi and Du, Chao and Pang, Tianyu and Liu, Qian and Zeng, Guangtao and Lin, Min and Li, Chongxuan},
  journal={arXiv preprint arXiv:2410.18514},
  year={2024}
}

@article{gong2024scaling,
  title={Scaling Diffusion Language Models via Adaptation from Autoregressive Models},
  author={Gong, Shansan and Agarwal, Shivam and Zhang, Yizhe and Ye, Jiacheng and Zheng, Lin and Li, Mukai and An, Chenxin and Zhao, Peilin and Bi, Wei and Han, Jiawei and others},
  journal={arXiv preprint arXiv:2410.17891},
  year={2024}
}

@article{ni2025training,
  title={Training optimal large diffusion language models},
  author={Ni, Jinjie and Liu, Qian and Du, Chao and Dou, Longxu and Yan, Hang and Wang, Zili and Pang, Tianyu and Shieh, Michael Qizhe},
  journal={arXiv preprint arXiv:2510.03280},
  year={2025}
}

@online{OpenAI2024o1,
  author    = {OpenAI},
  title     = {O1: OpenAI's First Model},
  year      = {2024},
  url       = {https://openai.com/o1/},
  note      = {Accessed: 2024-12-25}
}

@online{deepseekv3,
  author    = {DeepSeek-AI},
  title     = {DeepSeek-V3 Technical Report},
  year      = {2024},
  url       = {https://github.com/deepseek-ai/DeepSeek-V3/blob/main/DeepSeek_V3.pdf},
  note      = {Accessed: 2024-12-26}
}

@article{schulman2017proximal,
  title={Proximal policy optimization algorithms},
  author={Schulman, John and Wolski, Filip and Dhariwal, Prafulla and Radford, Alec and Klimov, Oleg},
  journal={arXiv preprint arXiv:1707.06347},
  year={2017}
}

@article{guo2025deepseek,
  title={Deepseek-r1: Incentivizing reasoning capability in llms via reinforcement learning},
  author={Guo, Daya and Yang, Dejian and Zhang, Haowei and Song, Junxiao and Zhang, Ruoyu and Xu, Runxin and Zhu, Qihao and Ma, Shirong and Wang, Peiyi and Bi, Xiao and others},
  journal={arXiv preprint arXiv:2501.12948},
  year={2025}
}

@inproceedings{rajbhandari2020zero,
  author       = {Samyam Rajbhandari and
                  Jeff Rasley and
                  Olatunji Ruwase and
                  Yuxiong He},
  editor       = {Christine Cuicchi and
                  Irene Qualters and
                  William T. Kramer},
  title        = {ZeRO: memory optimizations toward training trillion parameter models},
  booktitle    = {Proceedings of the International Conference for High Performance Computing,
                  Networking, Storage and Analysis, {SC} 2020, Virtual Event / Atlanta,
                  Georgia, USA, November 9-19, 2020},
  pages        = {20},
  publisher    = {{IEEE/ACM}},
  year         = {2020},
  url          = {https://doi.org/10.1109/SC41405.2020.00024},
  doi          = {10.1109/SC41405.2020.00024},
  bibsource    = {dblp computer science bibliography, https://dblp.org}
}

@inproceedings{dao2022flashattention,
  author       = {Tri Dao and
                  Daniel Y. Fu and
                  Stefano Ermon and
                  Atri Rudra and
                  Christopher R{\'{e}}},
  editor       = {Sanmi Koyejo and
                  S. Mohamed and
                  A. Agarwal and
                  Danielle Belgrave and
                  K. Cho and
                  A. Oh},
  title        = {FlashAttention: Fast and Memory-Efficient Exact Attention with IO-Awareness},
  booktitle    = {Advances in Neural Information Processing Systems 35: Annual Conference
                  on Neural Information Processing Systems 2022, NeurIPS 2022, New Orleans,
                  LA, USA, November 28 - December 9, 2022},
  year         = {2022},
  url          = {http://papers.nips.cc/paper\_files/paper/2022/hash/67d57c32e20fd0a7a302cb81d36e40d5-Abstract-Conference.html},
  bibsource    = {dblp computer science bibliography, https://dblp.org}
}

@article{dao2023flashattention,
  author       = {Tri Dao},
  title        = {FlashAttention-2: Faster Attention with Better Parallelism and Work
                  Partitioning},
  journal      = {CoRR},
  volume       = {abs/2307.08691},
  year         = {2023},
  url          = {https://doi.org/10.48550/arXiv.2307.08691},
  doi          = {10.48550/ARXIV.2307.08691},
  eprinttype    = {arXiv},
  eprint       = {2307.08691},
  bibsource    = {dblp computer science bibliography, https://dblp.org}
}

@article{dong2024flex,
  title={Flex Attention: A Programming Model for Generating Optimized Attention Kernels},
  author={Dong, Juechu and Feng, Boyuan and Guessous, Driss and Liang, Yanbo and He, Horace},
  journal={arXiv preprint arXiv:2412.05496},
  year={2024}
}

@article{zheng2024llamafactory,
  title={Llamafactory: Unified efficient fine-tuning of 100+ language models},
  author={Zheng, Yaowei and Zhang, Richong and Zhang, Junhao and Ye, Yanhan and Luo, Zheyan and Feng, Zhangchi and Ma, Yongqiang},
  journal={arXiv preprint arXiv:2403.13372},
  year={2024}
}

@misc{2023lmdeploy,
    title={LMDeploy: A Toolkit for Compressing, Deploying, and Serving LLM},
    author={InternLM},
    howpublished = {\url{https://github.com/InternLM/lmdeploy}},
    year={2023}
}

@article{zeng2025glm,
  title={Glm-4.5: Agentic, reasoning, and coding (arc) foundation models},
  author={Zeng, Aohan and Lv, Xin and Zheng, Qinkai and Hou, Zhenyu and Chen, Bin and Xie, Chengxing and Wang, Cunxiang and Yin, Da and Zeng, Hao and Zhang, Jiajie and others},
  journal={arXiv preprint arXiv:2508.06471},
  year={2025}
}

@misc{glm46,
    title={GLM-4.6: Advanced Agentic, Reasoning and Coding Capabilities},
    author={zai-org},
    howpublished = {\url{https://z.ai/blog/glm-4.6}},
    year={2025}
}

@misc{yu2025dapoopensourcellmreinforcement,
      title={DAPO: An Open-Source LLM Reinforcement Learning System at Scale}, 
      author={Qiying Yu and Zheng Zhang and Ruofei Zhu and Yufeng Yuan and Xiaochen Zuo and Yu Yue and Weinan Dai and Tiantian Fan and Gaohong Liu and Lingjun Liu and Xin Liu and Haibin Lin and Zhiqi Lin and Bole Ma and Guangming Sheng and Yuxuan Tong and Chi Zhang and Mofan Zhang and Wang Zhang and Hang Zhu and Jinhua Zhu and Jiaze Chen and Jiangjie Chen and Chengyi Wang and Hongli Yu and Yuxuan Song and Xiangpeng Wei and Hao Zhou and Jingjing Liu and Wei-Ying Ma and Ya-Qin Zhang and Lin Yan and Mu Qiao and Yonghui Wu and Mingxuan Wang},
      year={2025},
      eprint={2503.14476},
      archivePrefix={arXiv},
      primaryClass={cs.LG},
      url={https://arxiv.org/abs/2503.14476}, 
}

@article{qwen2024qwen25technicalreport,
      title={Qwen2.5 Technical Report}, 
      author={Qwen and : and An Yang and Baosong Yang and Beichen Zhang and Binyuan Hui and Bo Zheng and Bowen Yu and Chengyuan Li and Dayiheng Liu and Fei Huang and Haoran Wei and Huan Lin and Jian Yang and Jianhong Tu and Jianwei Zhang and Jianxin Yang and Jiaxi Yang and Jingren Zhou and Junyang Lin and Kai Dang and Keming Lu and Keqin Bao and Kexin Yang and Le Yu and Mei Li and Mingfeng Xue and Pei Zhang and Qin Zhu and Rui Men and Runji Lin and Tianhao Li and Tingyu Xia and Xingzhang Ren and Xuancheng Ren and Yang Fan and Yang Su and Yichang Zhang and Yu Wan and Yuqiong Liu and Zeyu Cui and Zhenru Zhang and Zihan Qiu},
      year={2024},
      journal={arXiv preprint arXiv:2412.15115},
      url={https://arxiv.org/abs/2412.15115}, 
}

@article{cobbe2021gsm8k,
  title={Training verifiers to solve math word problems},
  author={Cobbe, Karl and Kosaraju, Vineet and Bavarian, Mohammad and Chen, Mark and Jun, Heewoo and Kaiser, Lukasz and Plappert, Matthias and Tworek, Jerry and Hilton, Jacob and Nakano, Reiichiro and others},
  journal={arXiv preprint arXiv:2110.14168},
  year={2021}
}

@article{hendrycks2021math,
  title={Measuring mathematical problem solving with the math dataset},
  author={Hendrycks, Dan and Burns, Collin and Kadavath, Saurav and Arora, Akul and Basart, Steven and Tang, Eric and Song, Dawn and Steinhardt, Jacob},
  journal={arXiv preprint arXiv:2103.03874},
  year={2021}
}

@misc{aime2024,
  title={American Invitational Mathematics Examination-AIME 2024},
  author={MAA},
  year={2024},
  url={https://huggingface.co/datasets/math-ai/aime24}
}

@misc{aime2025,
  title={American Invitational Mathematics Examination-AIME 2024},
  author={MAA},
  year={2025},
  url={https://huggingface.co/datasets/math-ai/aime25}
}

@inproceedings{he2024olympiadbench,
  title={Olympiadbench: A challenging benchmark for promoting agi with olympiad-level bilingual multimodal scientific problems},
  author={He, Chaoqun and Luo, Renjie and Bai, Yuzhuo and Hu, Shengding and Thai, Zhen and Shen, Junhao and Hu, Jinyi and Han, Xu and Huang, Yujie and Zhang, Yuxiang and others},
  booktitle={Proceedings of the 62nd Annual Meeting of the Association for Computational Linguistics (Volume 1: Long Papers)},
  pages={3828--3850},
  year={2024}
}

@misc{wang2025revolutionizingreinforcementlearningframework,
      title={Revolutionizing Reinforcement Learning Framework for Diffusion Large Language Models}, 
      author={Yinjie Wang and Ling Yang and Bowen Li and Ye Tian and Ke Shen and Mengdi Wang},
      year={2025},
      eprint={2509.06949},
      archivePrefix={arXiv},
      primaryClass={cs.CL},
      url={https://arxiv.org/abs/2509.06949}, 
}

@misc{zhao2025d1scalingreasoningdiffusion,
      title={d1: Scaling Reasoning in Diffusion Large Language Models via Reinforcement Learning}, 
      author={Siyan Zhao and Devaansh Gupta and Qinqing Zheng and Aditya Grover},
      year={2025},
      eprint={2504.12216},
      archivePrefix={arXiv},
      primaryClass={cs.CL},
      url={https://arxiv.org/abs/2504.12216}, 
}

@misc{gong2025diffucoderunderstandingimprovingmasked,
      title={DiffuCoder: Understanding and Improving Masked Diffusion Models for Code Generation}, 
      author={Shansan Gong and Ruixiang Zhang and Huangjie Zheng and Jiatao Gu and Navdeep Jaitly and Lingpeng Kong and Yizhe Zhang},
      year={2025},
      eprint={2506.20639},
      archivePrefix={arXiv},
      primaryClass={cs.CL},
      url={https://arxiv.org/abs/2506.20639}, 
}

@misc{zhu2025vrporethinkingvaluemodeling,
      title={VRPO: Rethinking Value Modeling for Robust RL Training under Noisy Supervision}, 
      author={Dingwei Zhu and Shihan Dou and Zhiheng Xi and Senjie Jin and Guoqiang Zhang and Jiazheng Zhang and Junjie Ye and Mingxu Chai and Enyu Zhou and Ming Zhang and Caishuang Huang and Yunke Zhang and Yuran Wang and Tao Gui},
      year={2025},
      eprint={2508.03058},
      archivePrefix={arXiv},
      primaryClass={cs.LG},
      url={https://arxiv.org/abs/2508.03058}, 
}

@misc{lin2025boundaryguidedpolicyoptimizationmemoryefficient,
      title={Boundary-Guided Policy Optimization for Memory-efficient RL of Diffusion Large Language Models}, 
      author={Nianyi Lin and Jiajie Zhang and Lei Hou and Juanzi Li},
      year={2025},
      eprint={2510.11683},
      archivePrefix={arXiv},
      primaryClass={cs.LG},
      url={https://arxiv.org/abs/2510.11683}, 
}

@misc{rojas2025improvingreasoningdiffusionlanguage,
      title={Improving Reasoning for Diffusion Language Models via Group Diffusion Policy Optimization}, 
      author={Kevin Rojas and Jiahe Lin and Kashif Rasul and Anderson Schneider and Yuriy Nevmyvaka and Molei Tao and Wei Deng},
      year={2025},
      eprint={2510.08554},
      archivePrefix={arXiv},
      primaryClass={cs.LG},
      url={https://arxiv.org/abs/2510.08554}, 
}

@misc{zhao2025inpaintingguidedpolicyoptimizationdiffusion,
      title={Inpainting-Guided Policy Optimization for Diffusion Large Language Models}, 
      author={Siyan Zhao and Mengchen Liu and Jing Huang and Miao Liu and Chenyu Wang and Bo Liu and Yuandong Tian and Guan Pang and Sean Bell and Aditya Grover and Feiyu Chen},
      year={2025},
      eprint={2509.10396},
      archivePrefix={arXiv},
      primaryClass={cs.LG},
      url={https://arxiv.org/abs/2509.10396}, 
}
